\documentclass[conference]{IEEEtran}

\usepackage{cite}

\ifCLASSINFOpdf
\usepackage[pdftex]{graphicx}
\graphicspath{{figures/}}
\else
\fi

\usepackage{amsmath}

\usepackage{url}

\usepackage{tabularx}
\usepackage{multirow}
\usepackage{graphicx}
\usepackage{subcaption}
\usepackage{float}
\usepackage{amssymb}
\usepackage{algorithm}
\usepackage{algpseudocode}
\usepackage{multicol}
\usepackage{multirow}

\makeatletter
\renewcommand{\ALG@beginalgorithmic}{\footnotesize}
\makeatother

\begin{document}
	\bstctlcite{IEEEexample:BSTcontrol}
	\linespread{0.965}

\title{Distributed Learning approaches for Automated Chest X-Ray Diagnosis}
\author{
				\IEEEauthorblockN{Edoardo Giacomello}
		\IEEEauthorblockA{Dipartimento di Elettronica,\\
			Informazione e Bioinformatica\\
			Politecnico di Milano\\
			edoardo.giacomello@polimi.it}
		\and	
		\IEEEauthorblockN{Michele Cataldo}
		\IEEEauthorblockA{Dipartimento di Elettronica,\\
			Informazione e Bioinformatica\\
			Politecnico di Milano\\
			michele.cataldo@mail.polimi.it}
		\and
		\IEEEauthorblockN{Daniele Loiacono}
		\IEEEauthorblockA{Dipartimento di Elettronica,\\
			Informazione e Bioinformatica\\
			Politecnico di Milano\\
daniele.loiacono@polimi.it}
		\and
		\IEEEauthorblockN{Pier Luca Lanzi}
		\IEEEauthorblockA{Dipartimento di Elettronica,\\
			Informazione e Bioinformatica\\
			Politecnico di Milano\\
			pierluca.lanzi@polimi.it}		
	}

	\IEEEoverridecommandlockouts
	\IEEEpubid{\begin{minipage}{\textwidth}\ \\[12pt]
			978-1-7281-1884-0/19/\$31.00 \copyright 2019 IEEE
	\end{minipage}}
\maketitle
	
\begin{abstract}
		Deep Learning has established in the latest years as a successful approach to address a great variety of tasks. Healthcare is one of the most promising field of application for Deep Learning approaches since it would allow to help clinicians to analyze patient data and perform diagnoses.
However, despite the vast amount of data collected every year in hospitals and other clinical institutes, privacy regulations on sensitive data - such as those related to health - pose a serious challenge to the application of these methods.
In this work, we focus on strategies to cope with privacy issues when a consortium of healthcare institutions needs to train machine learning models for identifying a particular disease, comparing the performances of two recent distributed learning approaches - Federated Learning and Split Learning - on the task of Automated Chest X-Ray Diagnosis.
In particular, in our analysis we investigated the impact of different data distributions in client data and the possible policies on the frequency of data exchange between the institutions.
 	\end{abstract}

\IEEEpeerreviewmaketitle

	\section{Introduction}
	\label{sec:introduction}
	Thanks to the recent developments in the last few years, Deep Learning has already been established as a successful approach to address a great variety of tasks.
Among the others, healthcare is perhaps one of the most interesting fields of application for Deep Learning approaches, as it promises of helping clinicians with the analysis of patient data and with
a more accurate diagnostic process.
While in other fields, where deep learning is commonly employed, the abundance of data can ease the application of these techniques in practice, data availability is a serious concern for healthcare.
In fact, despite the vast amount of data collected every year in hospitals and other clinical institutes, privacy regulations on sensitive data - such are those related to health - pose a challenge to the feasibility of any data-driven approach.
Creating and maintaining large public datasets of medical information is a costly process since the patients participating in the study have to consent to allow the use of their data. Moreover, health data can present different modes based on the population: incidence of diseases can differ significantly by demographics and localization.
The challenges to this kind of approach can be traced to the single institution that collects data since acquisition methods can be different even in the same organization (e.g., different machinery models to acquire patient imaging).
Our work focuses on coping with data privacy issues when a consortium of healthcare institutions needs to train machine learning models for identifying a particular disease.
While the standard approach relies on anonymized data, this is not always possible - e.g., genomic data - and it adds more cost to the data collection process in terms of resources and time. In such a scenario, the most convenient option for developing a machine learning system would be exchanging the information - and not the data - relevant to the task between the different institutions and performing the training locally. To this extent, in this paper, we compare two recent distributed learning paradigms that allow completing privacy-preserving training on the task of diagnosing multiple diseases in Chest X-Ray images.

The paper is organized as follows.
In section \ref{sec:related}, we first introduce the concept of distributed training, then, in section \ref{sec:problemdefinition} we provide details on the task of choice. In section \ref{sec:methodology} we introduce the two selected distributed paradigms. Lastly, in sections \ref{sec:expdesign} and \ref{sec:results} we describe our experiments with different distributed layouts and training approaches.
 	
	\section{Related Work}
	\label{sec:related}
	\label{chap:state_of_the_art}

Distributed Machine Learning is an area of Machine Learning that aims at \textit{scaling out} classical machine learning algorithms in order to train larger models using a cluster of machines - eventually hosting multiple GPUs -. Over the years, the amount and complexity of data that must be managed in deep learning tasks have increased exponentially, to the point that single machines could not train the most complex models in a reasonable amount of time. Our work instead focuses on another crucial requirement of distributed machine learning: the possibility to process data while preserving privacy between the participants. This requirement became increasingly relevant as machine learning solutions have been proposed to learn from private user data, such as healthcare-related data or data collected from users' mobile devices.

Due to the twofold nature of the requirements, recent Distributed Machine Learning solutions usually pertain to Machine Learning, Computation Theory, Mathematics, and Computer Security. 
Research of privacy-preserving computation can be traced back to the late 1970s with the introduction of Homomorphic Encryption \cite{homomorphicorigin}, a kind of encryption that allows applying non-polynomial operations directly on encrypted data. In the following years, other methods have been proposed, such as Secure Multi-Party Computation (MPC) \cite{smpc}, a computational model that focuses on privacy between the participants of the computation instead of an external threat, and Differential Privacy (DP) in 2006 \cite{differentialprivacy}, to secure statistical databases.

With the rapid growth of available data and the introduction of Deep Learning, many studies have been proposed to adopt the concept above in Distributed Deep Learning Systems (DDLS). Most notably, Google introduced in 2012 DistBelief, an approach to perform Stochastic Gradient Descent on a cluster of thousands of machines \cite{distbelief2012}. In 2016, they introduced the concept of Federated Learning (FL) to perform distributed optimization that has been used to predict text in smartphone keyboards \cite{federatedaiblog} by adopting an MPC approach \cite{secureaggregation}. In 2018, MIT researchers proposed Split Learning, a different distributed learning approach that specifically addresses the use of Deep Neural Networks in a Healthcare setting \cite{DBLP:journals/corr/abs-1812-00564}.

While many other approaches have been proposed in the literature, we focus on Federated Learning and Split Learning in our work. Federated Learning has already been studied in healthcare settings, and previous results show that it can achieve performances that are very close to a centralized solution \cite{dlwsdata}. Split Learning, on the other hand, has been designed for health, and it can provide significant advantages in terms of versatility and privacy \cite{DBLP:journals/corr/abs-1812-00564, splitlearningcollaborativehealthcare}. To the best of our knowledge, no other works directly compare the two methods on the Automated Chest X-Ray diagnosis task.

\section{Automated Chest X-Ray Interpretation}
	\label{sec:problemdefinition}
	Automated Chest X-Ray Diagnosis is the task of training a machine learning model capable of predicting a series of diseases given an input Chest X-Ray (CXR) image. Our work aims to compare two distributed machine learning approaches, Split Learning and Federated Learning, in the context of Automated CXR Diagnosis. 
In this work, we used the CheXpert Dataset \cite{chexpert}, one of the most extensive datasets of labeled CXR images, released in 2019 by Stanford University to promote research into chest disease using X-Rays. The CheXpert dataset contains, to date, 224316 chest radiographs belonging to 65,240 patients, collected by the studies of Stanford Hospital in both inpatient and outpatient centers and performed between October 2002 and July 2017. The dataset is available online in two different versions: the first one provides high-quality images in 16-Bit PNG format; the second one stores images in 8-Bit PNG format. For our work, we have used the latter due to computational reasons. Every patient has at least a frontal CXR, while some patients also have a lateral image. Since the dataset is not complete in this sense, we only used frontal images in our experiments.
Each image is labeled with 14 different diseases. The labeling process was performed by an automated, rule-based labeler to extract observations from the free text radiology reports and involves a multi-step process that ultimately classifies each sample as Positive, Negative, or Uncertain.  
It is worth noting that the diseases are not mutually exclusive; this also due to the hierarchical nature of the diseases taken as reference in this study. For example, \textit{Atelectasis} is closely related to the presence of \textit{Lung Opacity}, or as \textit{Cardiomegaly} is a subcategory of \textit{Enlarged Cardiomediastinum}. Lastly, the authors of the dataset also provide a 202 image dataset to be used as the test set. In this case, the labeling is performed by radiology experts.
Before using the images with our models, we pre-processed each CXR by first resizing each image to 256x256, applying a template matching algorithm to find the area of interest for each image, and cropping the images to match the models' input size -i.e., 224x224-. The pixel values have been re-scaled in the range between 0 and 1, finally normalized using the ImageNet mean and standard deviation.
To cope with uncertain labels, we focused on the approaches proposed by Irvin et al. \cite{chexpert}. In particular, we considered the \emph{U-Ones} policy, which assigns the values of 1 to the uncertain labels. To prevent the model from becoming too much confident on these uncertain labels, we apply Label Smoothing Regularization \cite{DBLP:journals/corr/SzegedyVISW15}. Thus, we assign a randomly distributed value $x \sim U(a, b)$, with $a=0.55$ and $b=0.85$.
 	
	\section{Methodology}
	\label{sec:methodology}
	This section introduces the two different paradigms that we studied: Federated Learning and Split Learning. Both proposed methods have been designed as a solution for those situations in which a centralized approach is not feasible due to data privacy regulations, and a localized approach would require more data than is hosted by each client.

\subsection{Federated Learning}
\emph{Federated Learning} is a new distributed learning paradigm designed by Google \cite{mcmahan2017communicationefficient} in 2017. The key concept behind Federated Learning implementation is the decentralization of client models: the model is shared among the various users participating in the federated system while the data remains stored within each client, ensuring privacy and confidentiality.

Let us assume a scenario in which a federation of $n$ users $\{C_1, C_2, \cdots , C_n\}$ are coordinated by a central server to train a task-specific neural network. The workflow consists of 4 basic steps that are replicated over multiple federated rounds. 
\begin{enumerate}
	\setcounter{enumi}{-1}
	\item \textit{Initialization:} The server trains a generic neural network to initialize the configuration. \emph{This step is performed only in the first round}
	\item \textit{Clients selection:} A subset of clients is selected to participate in the current federated round. 
	\item \textit{Models updating:} The clients download the model from the server and train it based on their local data.
	\item \textit{Reporting:} All engaged clients send the updated model to the server (i.e., weights and bias of the local model).
	\item \textit{Aggregation:} The server aggregates the received parameters and updates the global model accordingly.
\end{enumerate}
The server now owns the updated model and is ready to start again from step 1 for another round. By iterating the process for multiple federated rounds, the global model is trained on client data without direct access to it. 

It is useful to emphasize that weights aggregation, from a server perspective, can be done in two different ways. The first one, called \emph{Federated Stochastic Gradient Descent (FedSGD)}, is the distributed implementation of the standard SGD algorithm. It is possible to implement a FedSGD by selecting the fraction $C$ of clients participating in a training round and a fixed learning rate $\eta$. In this way, each client \textit{k} compute $g_k=\triangledown F_k(w_t)$, the average gradients on its local data based on global model weights $w$ at the time $t$. The server at this point can proceed to update the weights by considering the proportion $\frac{n_k}{n}$ of samples from clients participating in the round compared to the total samples from all clients using the formula: $ w_{t+1} \leftarrow w_t - \eta \sum_{k=1}^K \frac{n_k}{n}g_k $
The second algorithm used for weights aggregation is \emph{Federated Averaging (FedAVG)}, a generalization of the previous one. It simply consists of updating the local model several times in each client \textit{k} before sending the weights to the server. This procedure can be helpful because, in FedSDG, if the local clients start from the same initialization, averaging the gradients is strictly equivalent to averaging the weights themselves.

An example of Federated Learning is shown in \cite{mcmahan2017communicationefficient}. The goal of this Federated system was to improve google's Gboard predictive keyboard based on data and user experience; hence, the privacy and confidentiality of texts written by users via the keyboard are the focus of the project. The system's clients are the users who have the GBoard keyboard installed on their mobile phones; instead, the server is an external entity on which the global model is located.
In the beginning, the server initializes a generic model in order to start the first federated round. Afterward, a subset of clients that are available for the training is selected. Then, each client device downloads the global model from the server (\emph{Clients Selection}) and trains the model on their data as input - i.e., the text written by the user - and updates the local model (\emph{Models Updating}). At this point, all clients send their modified model (\emph{Reporting}) to the server, which aggregates them and creates a new representation of the global model (\emph{Aggregation}). The new model will be sent to the clients during the next round.

\subsection{Split Learning}
\label{sec:split_learning}

The second approach we used is Split Learning, a paradigm developed in 2018 by MIT Media lab's Camera Culture Group \cite{DBLP:journals/corr/abs-1812-00564} that allows participating entities to train neural networks jointly, without sharing any raw data. Like Federated Learning, Split Learning has been developed to address model training when different institutions dealing with the same task have their collected data available, but these are insufficient to train a neural network capable of performing well. However, another interesting use case of Split Learning is when different entities or institutions hold different patient data modalities, such as electronic health records (EHR), pathological findings, and imaging data. These data taken individually are not suitable to train machine learning models; however, combining them would involve exchanging sensitive patient data, which is not always possible in healthcare due to privacy concerns. The Split Learning method for diagnosis would enable each center in this setting to contribute to constructing an aggregated model without sharing any raw data.

The key idea behind the Split Learning algorithm is that each entity (client) participating in the distributed system holds only a portion of the neural network model, from the top of the network until a specific \textit{Cut Layer}. During the forward pass, the output from the cut layer is sent to a third entity (server) that holds the remaining network section, which is univocal among all the clients. During the backward pass, the server can compute the gradients of its layers up to the cut layer. Then, the gradients of the cut layer are sent back to the clients, from which they can compute the remaining gradients up to the input layer. The peculiarity of this architecture is that the server split of the network is trained upon all the inputs received from the different clients. In this way, information can be shared among the client with no access to raw data.

This process just introduced can vary slightly depending on the architecture - or layout - that is chosen. It is possible to design a great variety of different architectures. In this paper, we focus on:
\subsubsection*{Vanilla architecture}
Figure \ref{fig:sl_vanilla} shows the most straightforward possible architecture of Split Learning, dubbed \textit{vanilla} by its authors, that corresponds to the process described previously. This architecture is applicable only if the server can access the labels of the samples to calculate gradients based on the loss function. 
More formally, assuming to have a neural network $F$ having $N$ layers $\{L_0, L_1, \cdots, L_N\}$, it can be \emph{split} in a way that there are \textit{local} layers, i.e. localized and accessed only by the client/institution, or \textit{shared} layers, hosted by a central server/institution. Assuming $L_m$ is the cut layer, with $0 < m < N$, the network will be split in $\{L_0, \cdots, L_m\} \in F_{client}$ local layers and $\{L_{m+1}, \cdots, L_N \} \in F_{server}$ and shared layers. During the training phase, the output $\hat{y}_m$ of $F_{client}$ is sent as an input to $F_{server}$, which completes the training obtaining the final predicted output $\hat{y}$. At this time, the server calculates the gradients $\mathbf{\rho}$, obtained by the loss function $G(\hat{y}, y)$, which are sent back through $F_{server}$ and $F_{client}$ for the back propagation.

\subsubsection*{U-Shaped architecture}
\begin{figure}[ht]
	\centering
	\begin{subfigure}{0.45\columnwidth}
		\centering
		\includegraphics[width=\textwidth]{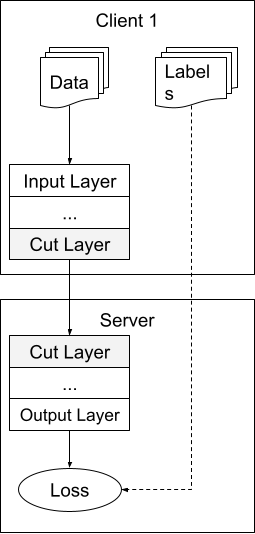}
		\subcaption
		{
			Vanilla Split Learning.
		}
		\label{fig:sl_vanilla}
	\end{subfigure}
	\begin{subfigure}{0.45\columnwidth}
		\centering
		\includegraphics[width=\textwidth]{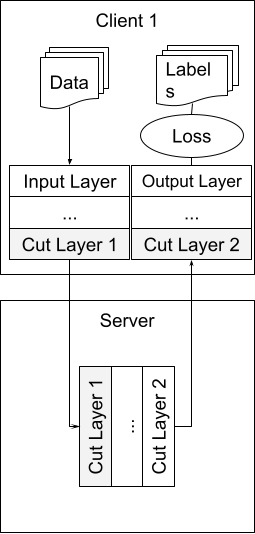}
		\subcaption
		{
			U-Shaped Split Learning.
		}
		\label{fig:sl_ushaped}
	\end{subfigure}
	\hfill
\caption[Split Learning Configurations]{Split Learning configurations.}
\end{figure}

The Vanilla configuration does not ensure full privacy when the labels contain sensitive data that should not be shared with the server. To overcome such a problem, adopting the U-Shaped configuration allows clients to preserve the labels, ensuring more privacy. In this case, as shown in Figure \ref{fig:sl_ushaped}, there is a third part of the network $F_{back}$, located in the client and composed of layers $\{L_{n+1}, \cdots, L_{N}\}$, restricting  consequently $F_{server}$ only to layers $\{L_{m+1}, \cdots, L_{n}\}$. The client will take care of the computation of $\hat{y}$ and gradients $\rho$ after receiving the intermediate output $\hat{y}_n$ from the server. This way, gradients will be sent back during backpropagation, first through the server layers and then into the client layers for weights updating. An important advantage of this architecture is that it allows each client to define its label semantic or use a different label set from the other clients.

	\section{Experimental Design}
	\label{sec:expdesign}
	\label{sec:experimental_design}

Our work aims to evaluate the performance of Federated Learning and Split Learning approaches in a distributed environment. 
We first designed an experiment where a centralized neural network is trained to compare the two methods more broadly. This experiment will be used as a benchmark during the evaluation of the models. The training of this network was performed with the same architectural settings as the distributed scenarios. For this first experiment, we exploited the entire CheXpert dataset using a ratio of 80\% - 20\% for the training/validation dataset selection. The resulting training dataset comprises $152.781$ samples, while the validation dataset is composed of $38.246$.

Focusing on the two distributed methods experiments, we designed a distributed environment where five clients and one server participate. We have considered all fourteen pathologies during the training phase. For the evaluation, we focused on the five most relevant pathologies (Atelectasis, Cardiomegaly, Consolidation, Edema, Pleural Effusion) in terms of clinical relevance and amount of samples. Each client is trained on different data from the others, while we used the same dataset for validation and testing phases.
The first dimension we analyzed during the experimental campaign is the distribution of labels in the client datasets. We have partitioned the training dataset following two different ways for assigning the respective datasets to the five clients. The first is a balanced approach, in which the distribution of pathologies across them is uniform. The average number of samples for the five balanced datasets is $35645 \pm 51$.
The second partitioning of data was done specifically to create a distribution of the five target pathologies such that each client owns a high presence of samples with a specific pathology. In particular, we ensured, whenever possible, that each client holds at least 45\% samples of a specific pathology.
The resulting unbalanced datasets  - and the corresponding clients - have been dubbed according to their majority label, resulting in the distribution: Consolidation ($57.2\%$), Cardiomegaly ($53.8\%$), Atelectasis ($51.3\%$), Edema ($45.8\%$), Pleural Effusion ($34.9\%$). It was impossible to reach the threshold of 45\% in the fifth client (Pleural Effusion) because of the high presence of this in samples that were also affected by other diseases. However, the dataset of the "Pleural Effusion" client is still consistent with our requirements since the client that comes closest in terms of the presence of Pleural Effusion samples is Client 4, with 14.328 samples. The other clients hold circa 10.000 Pleural Effusion samples. All the clients except Edema hold on average $35217.5 \pm 292$ samples, with the latter holding $37355$ images. The validation dataset used in both scenarios is composed of 12.802 samples.

The second dimension to take into account during our experiments is the granularity of data sharing between clients:
The \emph{fine grained} training of the neural network on each client occurs in small steps sequentially and synchronously: each client can train only on a single batch of data before the communication with the server. Then, the next client is selected to continue the training. The first client processes the second batch of data on the next iteration, and the process goes on until the dataset is exhausted. 
Conversely, we define as \emph{coarse grained} a distributed strategy in which every client performs the training on its entire dataset before the training can be performed on another client. 

As the last step to define our experimental campaign, we considered the case where clients work locally without data sharing to test the centralized learning capabilities when they cannot participate in a distributed system. For this experiment, we trained five client models using the unbalanced datasets.

\subsection{Implementation and Training}
Due to memory limitations, we could not use a large neural network architecture such as DenseNet121 with five clients. Instead, we decided to use a MobileNet architecture \cite{howard2017mobilenets}, which is still feasible for both Federated and Split Learning, even though the final performances have been negatively affected. On the other hand, the focus of this work is to study the differences with different distributed learning settings rather than finding the best possible model, so we leave experiments involving a more powerful architecture as future work.
Instead of training a Neural Network from scratch, we exploited Transfer Learning using a pre-trained Neural Network, so network weights were initialized with those of a MobileNet trained on the ImageNet dataset. At the end of the neural network, we added a Global Average Pooling followed by a new Fully Connected layer, which produces a 14-dimensional output, useful for our problem, after which we have applied a sigmoid activation function. 

Since we wanted to benchmark the model performances using each paradigm, we did not simulate a real case scenario where every client physically resides on a different machine. Instead, we implemented the distributed systems on a single instance using TensorFlow 2.5 and Google Colab. We implemented Federated Learning using the Tensorflow Federated module, which is currently in development. For implementing the coarse-grained strategy, we select all the clients during each training phase. To implement fine-grained granularity in Federated Learning, we needed to create multiple virtual clients for each actual client, each one holding a single batch of data. Accordingly, we modified the client selection process and the number of rounds so that at every federated round, each client trained its model on one batch of data, ensuring no overlap in training batches. 
Due to software limitations, for implementing Split Learning, we needed to simulate the server's behavior. In particular, each client owns the entire neural network, and the weights corresponding to the server split are copied on the other clients' model whenever a server update occurs. With the fine-grained strategy, the server weights are broadcasted to the clients after each client trains a single batch of data. After that, the next client processes its batch of data. Conversely, in the coarse-grained strategy, the switch of each client is performed after the previous one has completed the training on the whole dataset.
The network has been split at the 60th layer for the Vanilla architecture, leaving $66\%$ of the weights on server and $33\%$ on each client. The choice for the second cut in the U-Shaped architecture has been limited by the weights amount distribution across the network layers. To promote a fair comparison between Vanilla and U-Shaped architecture, we choose to cut the network just before the last Fully Connected Layer, which is sufficient to avoid disclosing patient labels while keeping the ratio between client and server weights almost unaltered. 

For training the networks on all the experiments, we used a Stochastic Gradient Descent optimizer with a learning rate of $1e-3$. The number of epochs has been set to a maximum of 10, using Early Stopping with a \emph{patience} factor of 4. The loss function is the binary cross-entropy loss between the ground truth labels and the outputs. The Federated Learning model training was performed using the FedSGD algorithm because it is the dual implementation of the SGD optimizer used in the training of client models in the Split Learning approach. The learning rate of clients is $1e-3$, while the server learning rate has the default value of 1. It is important to note that the learning rate of the server is used during the averaging process and therefore differs from the traditional role of a generic learning rate in a centralized system. 	
	\section{Results}
	\label{sec:results}
	\begin{table*}[htb]
	\centering
\begin{tabular}{|l|c|c|c|c|c|c|} 
	\hline
	{} &  Atelectasis &  Cardiomegaly &  Consolidation &  Edema &  Pleural Effusion &   Mean \\ \hline
	Centralized                                  &       0.7675 &        0.7135 &         0.8285 & 0.7890 &            0.8329 & 0.7863 \\  \hline
	Federated Learning, Uniform, Fine            &       0.7604 &        0.6416 &         0.7406 & 0.8549 &            0.8560 & 0.7707 \\  \hline
	Federated Learning, Uniform, Coarse          &       0.7165 &        0.7237 &         0.7208 & 0.8460 &            0.8384 & 0.7691 \\  \hline
	Split Learning, Uniform, Vanilla, Fine       &       0.6867 &        0.7086 &         0.7366 & 0.8283 &            0.7528 & 0.7426 \\  \hline
	Split Learning, Uniform, Vanilla, Coarse     &       0.6898 &        0.6991 &         0.7426 & 0.8256 &            0.7511 & 0.7416 \\  \hline
	Split Learning, Uniform, U-Shaped, Fine      &       0.7366 &        0.6507 &         0.6301 & 0.8174 &            0.8074 & 0.7285 \\  \hline
	Split Learning, Uniform, U-Shaped, Coarse    &       0.7094 &        0.6451 &         0.6494 & 0.7954 &            0.8057 & 0.7210 \\  \hline
	Local, Uniform, Client 0                     &       0.6207 &        0.6432 &         0.7184 & 0.7164 &            0.6363 & 0.6670 \\  \hline
	Local, Uniform, Client 1                     &       0.6213 &        0.6393 &         0.7033 & 0.7156 &            0.5918 & 0.6543 \\  \hline
	Local, Uniform, Client 2                     &       0.5929 &        0.6472 &         0.7017 & 0.6984 &            0.6483 & 0.6577 \\  \hline
	Local, Uniform, Client 3                     &       0.6275 &        0.6329 &         0.6800 & 0.6958 &            0.6150 & 0.6502 \\  \hline
	Local, Uniform, Client 4                     &       0.6531 &        0.6379 &         0.6998 & 0.7174 &            0.6555 & 0.6727 \\  \hline
	Federated Learning, Unbalanced, Fine         &       0.6413 &        0.7147 &         0.7057 & 0.8876 &            0.8651 & 0.7629 \\  \hline
	Federated Learning, Unbalanced, Coarse       &       0.7910 &        0.6829 &         0.7243 & 0.8134 &            0.8145 & 0.7652 \\  \hline
	Split Learning, Unbalanced, Vanilla, Fine    &       0.7160 &        0.6805 &         0.7197 & 0.7868 &            0.7372 & 0.7280 \\  \hline
	Split Learning, Unbalanced, Vanilla, Coarse  &       0.7179 &        0.6902 &         0.6960 & 0.8077 &            0.7362 & 0.7296 \\  \hline
	Split Learning, Unbalanced, U-Shaped, Fine   &       0.7512 &        0.5948 &         0.6783 & 0.7580 &            0.8124 & 0.7189 \\  \hline
	Split Learning, Unbalanced, U-Shaped, Coarse &       0.7504 &        0.5922 &         0.6478 & 0.7542 &            0.8040 & 0.7097 \\  \hline
	Local, Unbalanced, atel                      &       0.6655 &        0.6287 &         0.6706 & 0.7086 &            0.6303 & 0.6607 \\  \hline
	Local, Unbalanced, card                      &       0.5710 &        0.5954 &         0.6493 & 0.6124 &            0.4899 & 0.5836 \\  \hline 
	Local, Unbalanced, cons                      &       0.6511 &        0.6185 &         0.6969 & 0.6561 &            0.6166 & 0.6479 \\  \hline
	Local, Unbalanced, edema                     &       0.6154 &        0.6205 &         0.6105 & 0.7326 &            0.6098 & 0.6378 \\  \hline
	Local, Unbalanced, peff                      &       0.6094 &        0.6140 &         0.6732 & 0.7058 &            0.6436 & 0.6492 \\  \hline
\end{tabular}
    \caption{Individual label scores of the average model obtained in each experiment. The name of the experiment reports the learning paradigm, the dataset distribution (uniform or unbalanced) and the data granularity (Fine or Grain). For Split learning is also indicated the architecture layout (Vanilla or U-Shaped).}
	\label{tab:label_scores}
\end{table*}

In this section, we show and discuss our experimental results. First, in Table \ref{tab:label_scores}, we present our results on a Centralized model, then we evaluate each proposed learning paradigm first on a uniform dataset distribution and then on an unbalanced one. For these analyses, we compared the average model obtained with each method -i.e., each Split Learning score is the average prediction over all the resulting client models -. Then, we proceed to evaluate the performances of the client model we obtained -Table \ref{tab:client_scores}-, where applicable. Lastly, in Table \ref{tab:summary_scores}, we propose an overview of the performances of the considered distributed paradigms. 

\subsection{Centralized Model}
To benchmark our results, we first trained a centralized model using all the available data. This centralized model achieves an average AUC of 0.7863, which is in line with our previous experiments using the same data and architecture. Analyzing the scores for the specific labels, the model has difficulty recognizing Cardiomegaly while performing well on Pleural Effusion and Consolidation, reaching an AUC of 0.83. 

\subsection{Uniform Datasets}
The first set of experiments on distributed learning paradigms have been carried on using uniform datasets.

The first distributed paradigm we used is Federated Learning with a coarse-grained data sharing policy. In the case of uniform datasets, this approach proved to work very well and achieves performances only slightly lower than the centralized model. The average AUC is 0.7691, reaching values around 0.85 for labels such as Edema and Pleural Effusion. 

The Federated Learning model, trained with fine-grained sharing data, has the best classification score, reaching performances similar to a centralized model while providing the advantages that a distributed approach can offer. More specifically, the model reaches an average AUC of 0.7707 and excellent results for the pathologies Edema and Pleural Effusion, exceeding in these cases the performance of the centralized model. However, the score on Cardiomegaly drops sharply to a value of 0.64, compared to the other two experiments discussed so far.

The first of the Split Learning experiments involves the training of a \textit{vanilla} architecture with coarse granularity and uniform datasets. Despite a decrease of the average AUROC that falls to 0.7416, the model is still a good classifier with AUCs for individual labels above the threshold of 0.70 and an AUC of 0.83 for Edema. Similarly, the corresponding model with fine granularity only led to a slight difference in the average AUC (0.7426 vs. 0.7416), and the AUC for individual labels is very close to those of the previous experiment.
In the following experiments, we used Split Learning with the U-Shaped architecture. When using a fine-grained sharing policy, we obtained a mean AUC of 0.7285, which is slightly lower than the \textit{vanilla} architecture using the same sharing policy but comes with the benefit of an increase in privacy between the client and server. The individual scores show a decrease in performance for Cardiomegaly (0.6507) and Consolidation (0.6301), while performances of Atelectasis (0.7366) and Pleural Effusion (0.8074) are better than the vanilla counterpart. Even for this architecture, the AUC scores are similar to the fine-grained policy when using a coarse-grained sharing policy (0.7210 vs. 0.7285). The comparison of individual scores between the two data-sharing policies follows roughly the same pattern seen for the vanilla architecture. In conclusion, our results on the Split Learning approach suggest that Split Learning is quite robust to changes in data sharing granularity.

To highlight the importance of introducing data sharing between clients, we performed an additional experiment where we disabled the communication between clients. In other words, every client hosts a model that is trained only on its dataset.
The best AUC is that of client 4 that manages to reach a maximum of 0.6727, while the other clients oscillate around an AUROC of 0.65. However, none of the 5 cases would represent a classifier good enough to be used in the medical practice because the risk of incorrect prediction is very high. The average of the AUCs of client models is equal to 0.6604.

\subsection{Unbalanced Datasets}
In this section, we show our results obtained by repeating the previous experiments on our unbalanced datasets. We unbalanced each client's dataset so that each client had a prevalence of samples of a particular disease.

Even in the case of unbalanced, the coarse-grained Federated Learning model performs very well, achieving an average AUC of 0.7652 and similar results to the centralized baseline model. Changing the dataset distribution among the clients showed a minimal effect on the performance of this distributed approach. Comparing the results with the corresponding results for the uniform dataset, we note that the AUC values for the Edema and Pleular Effusion labels are slightly decreased. However, an improvement in the Atelectasis score is also noteworthy, which increases from 0.72 to 0.79.
The model trained with Federated Learning and fine granularity achieves the highest AUC, 0.887, for Edema pathology, outperforming even the centralized model. Pleural Effusion also ranks well, presenting an AUC of 0.86. These two categories are confirmed to be among the easiest to recognize within this study. The mean AUC of the model is 0.7629, which allows it to rank among the best classifiers in our experiments. On the other hand, we must report difficulties in classifying Atelectasis correctly, with an AUC of 0.64. 

The results of Vanilla Split Learning with coarse granularity presents a similar scenario to the balanced dataset case, in which we observe a slight loss of performances compared to Federated Learning. However, the resulting classifier predicts well most of the diseases, such as Edema (0.81), Atelectasis (0.72), and Pleulal Effusion (0.74). Cardiomegaly and Consolidation pathologies remain the most difficult to classify, probably due to their lower representation in the dataset. Concerning the Split Learning model with fine-grained data sharing, no relevant differences are found to coarse-grained as this has an average AUC of 0.73. The best label is Edema, with an AUC of 0.79, while Cardiomegaly reaches 0.69. Recalling the same models trained with uniform datasets, we note that for both fine and coarse grain, the overall performance is slightly lower with unbalanced datasets, a sign that changing the distribution of client datasets affects, although only slightly, the AUC score. 

When using the U-Shaped architecture, we can observe the same behavior seen in the balanced dataset scenario: the average score for both Fine and Coarse granularity are slightly lower than those of the vanilla architecture - 0.710 vs. 0.730 for Coarse Granularity and 0.719 vs. 0.728 for Fine Granularity-. It is interesting to note that again, Pleural Effusion obtains a substantial boost in performances when choosing this architecture, while the scores on other labels such Cardiomegaly and Edema suffer from the change of the architecture. This may indicate that some labels may be more sensitive than others to changes in the last classification layer and deserve more insights.

\begin{table*}[htb]
	\centering
	\begin{tabular}{|l|l|c|c|c|c|c|c|}
		\hline
		Client &                 Paradigm &  Atelectasis &  Cardiomegaly &  Consolidation &  Edema &  Pleural Effusion &   Mean \\ \hline \hline
		Atelectasis      &                    Local &       0.6655 &        0.6287 &         0.6706 & 0.7086 &            0.6303 & 0.6607 \\  \hline
		Atelectasis      &  Split, U-Shaped, Coarse &       0.6408 &        0.5624 &         0.5221 & 0.6333 &            \textbf{0.7537} & 0.6225 \\  \hline
		Atelectasis      &    Split, U-Shaped, Fine &       0.6206 &        0.5418 &         0.5408 & 0.6732 &            0.7258 & 0.6204 \\  \hline
		Atelectasis      &   Split, Vanilla, Coarse &       0.6871 &        \textbf{0.6911} &         0.6643 & \textbf{0.8073 }&            0.7458 & \textbf{0.7191} \\  \hline
		Atelectasis      &     Split, Vanilla, Fine &       \textbf{0.6940} &        0.6612 &         \textbf{0.6991} & 0.7638 &            0.7272 & 0.7091 \\  \hline \hline
		
		Cardiomegaly     &                    Local &       0.5710 &        0.5954 &         0.6493 & 0.6124 &            0.4899 & 0.5836 \\  \hline
		Cardiomegaly     &  Split, U-Shaped, Coarse &       0.6390 &        0.4617 &         0.5379 & 0.7475 &            0.7631 & 0.6298 \\  \hline
		Cardiomegaly     &    Split, U-Shaped, Fine &       0.6341 &        0.4678 &         0.5425 & 0.7604 &            \textbf{0.7685} & 0.6347 \\  \hline
		Cardiomegaly     &   Split, Vanilla, Coarse &       \textbf{0.7447} &        \textbf{0.6834} &         \textbf{0.6805} & \textbf{0.7888} &            0.7370 & 0.7269 \\  \hline
		Cardiomegaly     &     Split, Vanilla, Fine &       0.7231 &        0.6562 &         0.7346 & 0.7874 &            0.7344 & \textbf{0.7271} \\  \hline \hline
		
		Consolidation    &                    Local &       0.6511 &        0.6185 &         0.6969 & 0.6561 &            0.6166 & 0.6479 \\  \hline
		Consolidation    &  Split, U-Shaped, Coarse &       0.6833 &        0.6535 &         0.6579 & 0.7275 &            0.6558 & 0.6756 \\  \hline
		Consolidation    &    Split, U-Shaped, Fine &       0.6693 &        0.6631 &         0.6882 & 0.7272 &            0.6598 & 0.6815 \\  \hline
		Consolidation    &   Split, Vanilla, Coarse &       0.6998 &        \textbf{0.6857} &         0.7026 & \textbf{0.7865} &            0.7057 & 0.7161 \\  \hline
		Consolidation    &     Split, Vanilla, Fine &       \textbf{0.7019} &        0.6786 &         \textbf{0.7040} & 0.7807 &            \textbf{0.7217 }& \textbf{0.7174 } \\  \hline \hline
		
		Edema            &                    Local &       0.6154 &        0.6205 &         0.6105 & 0.7326 &            0.6098 & 0.6378 \\  \hline
		Edema            &  Split, U-Shaped, Coarse &       0.6331 &        0.5430 &         0.6371 & 0.5228 &            0.7509 & 0.6174 \\  \hline
		Edema            &    Split, U-Shaped, Fine &       0.6417 &        0.5447 &         0.6577 & 0.5301 &            \textbf{0.7574} & 0.6263 \\  \hline
		Edema            &   Split, Vanilla, Coarse &       0.6966 &        0.6722 &         0.6649 & \textbf{0.8062} &            0.7146 & 0.7109 \\  \hline
		Edema            &     Split, Vanilla, Fine &       \textbf{0.7124} &        \textbf{0.6950} &         \textbf{0.6676} & 0.7802 &            0.7263 & \textbf{0.7163} \\  \hline \hline
		
		Pleural Effusion &                    Local &       0.6094 &        0.6140 &         0.6732 & 0.7058 &            0.6436 & 0.6492 \\  \hline
		Pleural Effusion &  Split, U-Shaped, Coarse &       0.5572 &        0.5264 &         0.5222 & 0.7049 &            0.6465 & 0.5914 \\  \hline
		Pleural Effusion &    Split, U-Shaped, Fine &       0.6168 &        0.5175 &         0.5188 & 0.6856 &            0.6658 & 0.6009 \\  \hline
		Pleural Effusion &   Split, Vanilla, Coarse &       0.6759 &        0.6492 &         0.6814 & \textbf{0.7653} &            0.7053 & 0.6954 \\  \hline
		Pleural Effusion &     Split, Vanilla, Fine &       \textbf{0.6951} &        \textbf{0.6629} &         \textbf{0.7112 }& 0.7528 &            \textbf{0.7101} & \textbf{0.7064} \\  \hline
	\end{tabular}
	\caption{Comparison of the client model for each label.}
	\label{tab:client_scores}
\end{table*}

To complete the picture of the experiments on unbalanced datasets, we again analyzed the behavior of client models without the possibility of collaboration. Table \ref{tab:label_scores} shows that the models has generally low performances. For instance, the client trained on a prevalence of Cardiomegaly samples achieves an average AUC of 0.58, slightly higher than that of a random classifier. The average results are lower than the uniform dataset scenario, which is not always available in reality. These scores represent the results that could be achieved in a real case scenario in which each client is specialized on a particular mode of data without applying distributed training. Having set our lower bound on performances, the benefit of distributed training becomes now more evident. 

As the last analysis, we compare the local models with the corresponding model trained using the different Split Learning methodologies. Table \ref{tab:client_scores} shows that, in every case, applying Split Learning resulted in being beneficial in terms of average AUC compared to the local client performances; this proves that Split Learning is indeed capable of transferring valuable information to the clients without disclosing input samples to the other participants. When considering individual label scores, the best results are obtained by applying Vanilla Split Learning with either coarse or fine granularity. Unsurprisingly, U-Shaped architectures perform slightly worse than the vanilla counterpart due to the last classification layer not being shared anymore. However, it can still be proven beneficial on some target labels - such as Pleural Effusion - or when the local clients perform poorly due to their local dataset distribution.

\subsection{Paradigm Comparison}
Table \ref{tab:summary_scores} summarizes the mean AUC scores for every experiment setup. As expected, the performance of the distributed approaches ranked between the upper benchmark, defined by the centralized model, and the lower benchmark, defined by the local models trained without data sharing.
The Federated Learning models, achieving AUCs roughly similar to the baseline model, performed better than the Split Learning models. However, the latter provides more versatility in the model's design, and they have achieved good overall performance.

The granularity of data sharing did not affect Federated Learning nor the Split Learning paradigms to a relevant degree. Regarding the choice of datasets, the Federated Learning models do not seem to have been affected by the modification, confirming once again to be good classifiers. Instead, using unbalanced datasets in Split Learning has slightly lowered the performance of the models, which found more difficulty in the classification of diseases but still obtaining acceptable results. We envision that, in practice, this loss of performance may be compensated by designing the client shard of the network more carefully, according to the needs dictated by the individual dataset distributions. On the other hand, this may be difficult to implement in Federated Learning as the architecture is fixed for all the clients.
Finally, we can point out that the easiest pathologies to classify are Edema and Pleural Effusion for almost all the models taken into account; these pathologies are also the most present in the training dataset, and for this reason, the models are more confident with their classification. Similarly, the pathology Cardiomegaly and occasionally Consolidation are the ones that in most models are more difficult to predict and may require a better choice of the architecture.

\begin{table}[htb]
	\centering
	\begin{tabular}{|c|c|c|c|}
		\hline
		\textbf{\begin{tabular}[c]{@{}c@{}}Learning\\ Paradigm\end{tabular}} &
		\textbf{\begin{tabular}[c]{@{}c@{}}Data Sharing\\ Granularity\end{tabular}} &
		\textbf{\begin{tabular}[c]{@{}c@{}}Uniform\\ Datasets\end{tabular}} &
		\textbf{\begin{tabular}[c]{@{}c@{}}Unbalanced\\ Datasets\end{tabular}} \\ \hline
		Centralized Model                  & \textbf{-} & \multicolumn{2}{c|}{{0.7863}} \\ \hline
		\multirow{2}{*}{Federated Learning} & Coarse     & 0.7691       & 0.7652       \\ \cline{2-4} 
		& Fine       & 0.7707       & 0.7629       \\ \hline
		\multirow{3}{*}{Split Learning (Vanilla)}     & Coarse     & {0.7416}       & 0.7296       \\ \cline{2-4} 
		& Fine       & 0.7426       & 0.7280       \\ \hline
		\multirow{3}{*}{Split Learning (U-Shaped)}     & Coarse  &  0.7210    &   0.7097  \\ \cline{2-4} 
		& Fine       &  0.7285   &   0.7189  \\ \hline
		Local Models                        & \textbf{-} & 0.6604       & 0.6263       \\ \hline
	\end{tabular}
	\caption{Summary of Mean AUC for each experiment.}
	\label{tab:summary_scores}
\end{table}

	\section{Conclusions}
	\label{sec:conclusions}
	Distributed Learning is a promising approach to enable the training of Machine Learning models on large datasets while ensuring privacy between the participants. This aspect is particularly relevant in healthcare, where the privacy of patients is regulated by law. Thus, we considered two of the most popular approaches for distributed training in healthcare: Federated Learning and Split Learning. The goals of this work are: (i) compare the performances of the two methods, (ii) understand the impact of the frequency of information-sharing between the clients, and (iii) investigate the impact of client data distribution on each case study.
To this purpose, we trained several MobileNet CNNs on the CheXpert dataset, a large public dataset of more than 200k Chest X-Ray images. To test our approaches, we considered the five most relevant diseases in the dataset: Atelectasis, Cardiomegaly, Consolidation, Edema, Pleural Effusion.
As a reference, we first trained a CNN on the entire dataset to assess the performance of a centralized approach where privacy is not attained.
Then, to simulate a distributed scenario, we split the available data into five subsets, representing different institutions holding approximately the same amount of data. In the first set of 5 splits, data is split uniformly, meaning that all clients hold the same relative amount of labels. In the second set, we artificially unbalanced each client dataset such that in each client, there is a prevalence of data corresponding to each of the target labels.
To understand the impact of information sharing frequency, we considered two different granularities: in the coarse granularity, the data is exchanged after the client trained its model on the whole dataset. In the fine granularity, the clients only train a batch of data before communicating with the server.
We then performed our experiments training the same MobileNet architecture with Federated Learning and Split Learning -- both in Vanilla and U-Shaped layouts -- using each combination of dataset balance and data granularity policy.
Lastly, we trained a model for each client data split without using any distributed learning approach to set our baseline.

Our results show that both Federated Learning and Split Learning allow to achieve performances close to the one achieved with centralized training.
In addition, as expected, both the methods perform better with balanced data distribution, although they achieve reasonable performances even with unbalanced datasets.
In all the experimental settings, distributed methods outperform models trained on the local dataset, suggesting that these methods are worthy of further investigation and might be employed in practice to deal with data privacy issues in machine learning applications in the healthcare field.
Our results also showed a minor impact on the performances concerning how frequently clients share information with the server, making it possible to limit the amount of networking resources required in practice by the methods studied.
At the same time, we believe that future works are necessary to investigate better how distributed learning methods can perform with a smaller amount of data -- as commonly happens in the healthcare applications --  and with more sophisticated deep neural network architectures with respect to the MobileNet used in this work for keeping the computational effort limited.

	\bibliographystyle{ieeetran}
\bibliography{bibliography}
	
\end{document}